\documentclass{article}

\PassOptionsToPackage{numbers,sort&compress}{natbib}
\usepackage[preprint]{neurips_2026}

\usepackage[utf8]{inputenc} 
\usepackage[T1]{fontenc}    
\usepackage{hyperref}       
\usepackage{url}            
\usepackage{booktabs}       
\usepackage{amsfonts}       
\usepackage{nicefrac}       
\usepackage{microtype}      
\usepackage{xcolor}         

\usepackage{amsmath}
\usepackage{enumitem}
\usepackage{graphicx}
\usepackage{multirow}
\usepackage{pgffor}

\title{FreeSpec: Training-Free Long Video Generation via Singular-Spectrum Reconstruction}

\author{
    Fangda Chen$^{1}$, 
    Shanshan Zhao$^{2}$, 
    Longrong Yang$^{3}$,
    Chuanfu Xu$^{1}$,
    Zhigang Luo$^{4}$,
    Long Lan$^{1}$\thanks{Corresponding author} \\
    $^{1}$College of Computer Science and Technology, National University of Defense Technology \\
    $^{2}$Alibaba International Digital Commerce \\
    $^{3}$Zhejiang University \\
    $^{4}$Xiangjiang Laboratory, Changsha, Hunan, China \\
    {\tt\small \{fdchen.nudt, xuchuanfu, long.lan\}@nudt.edu.cn} \\
    {\tt\small sshan.zhao00@gmail.com, longrongyang@zju.edu.cn, Zhigang\_luo@sina.com}
}

\begin{document}

\maketitle

\vspace{-2em}
\begin{abstract}
Video diffusion models perform well in short-video synthesis, but their training-free extension to long videos often suffers from content drift, temporal inconsistency, and over-smoothed dynamics. Existing methods improve temporal consistency by combining a global branch with a local branch, but they often further decompose appearance consistency and temporal dynamics within each branch using predefined criteria. This assignment is unreliable when appearance and action progression are tightly coupled, such as in camera motion and sequential motion. We analyze the video temporal extension issue from a singular-spectrum perspective and show that enlarged self-attention windows induce spectral concentration: spectral energy becomes dominated by a few low-rank singular directions, preserving coarse structure but suppressing high-rank spatial details and motion-rich temporal variations. To mitigate this problem, we propose \textbf{FreeSpec}, a training-free spectral reconstruction framework for long-video generation. FreeSpec decomposes global and local features with singular value decomposition, and uses the global branch as low-rank spectral guidance and the local branch as a high-rank reconstruction basis. This spectrum-level fusion avoids the rigid feature partitioning of previous decomposition rules, preserving long-range consistency while better retaining spatial details and temporal dynamics. Experiments on Wan2.1 and LTX-Video demonstrate that FreeSpec improves long-video generation, especially for temporal dynamics, while maintaining strong visual quality and temporal consistency. Project demo: \url{https://fdchen24.github.io/FreeSpec-Website/}.
\end{abstract}

\vspace{-2em}
\section{Introduction}

Video diffusion models~\cite{videoldm, lavie, videocrafter2, cogvideox, hunyuanvideo, ltx, wan, jointtuner, school1} have achieved remarkable progress in short-video generation, producing realistic appearances, coherent motion, and strong text alignment over dozens of frames. However, directly training long-video diffusion models remains prohibitively expensive~\cite{streamingt2v, vidu, self_forcing}, due to the demand for large-scale long-duration video-text data, increased memory consumption, and substantial computation for modeling extended temporal dependencies. Therefore, a practical alternative is to extend pretrained short-video models to long-range synthesis in a training-free manner. 

However, training-free long-video generation remains challenging because pretrained short-video models are optimized within a limited temporal receptive field and often suffer from content drift, temporal inconsistency, repetitive patterns, or visual blurring when extrapolated to longer sequences. Recent studies tackle these limitations by revising the inference protocols of pretrained short-video models~\cite{gen_l_video, freenoise, fifo, riflex, longdiff, freeloc, freelong, freelong++, freepca}. For example, Gen-L-Video~\cite{gen_l_video}, FreeNoise~\cite{freenoise}, and FIFO-Diffusion~\cite{fifo} extend the temporal horizon with overlapping segments, noise rescheduling, or queue-based denoising. RIFLEx~\cite{riflex}, LongDiff~\cite{longdiff}, and FreeLOC~\cite{freeloc} improve length extrapolation by modifying positional encoding or correcting context-length distribution shifts. 

\begin{figure*}[t]
    \centering
    \hspace*{-0.05\linewidth}
    \includegraphics[width=0.95\linewidth]{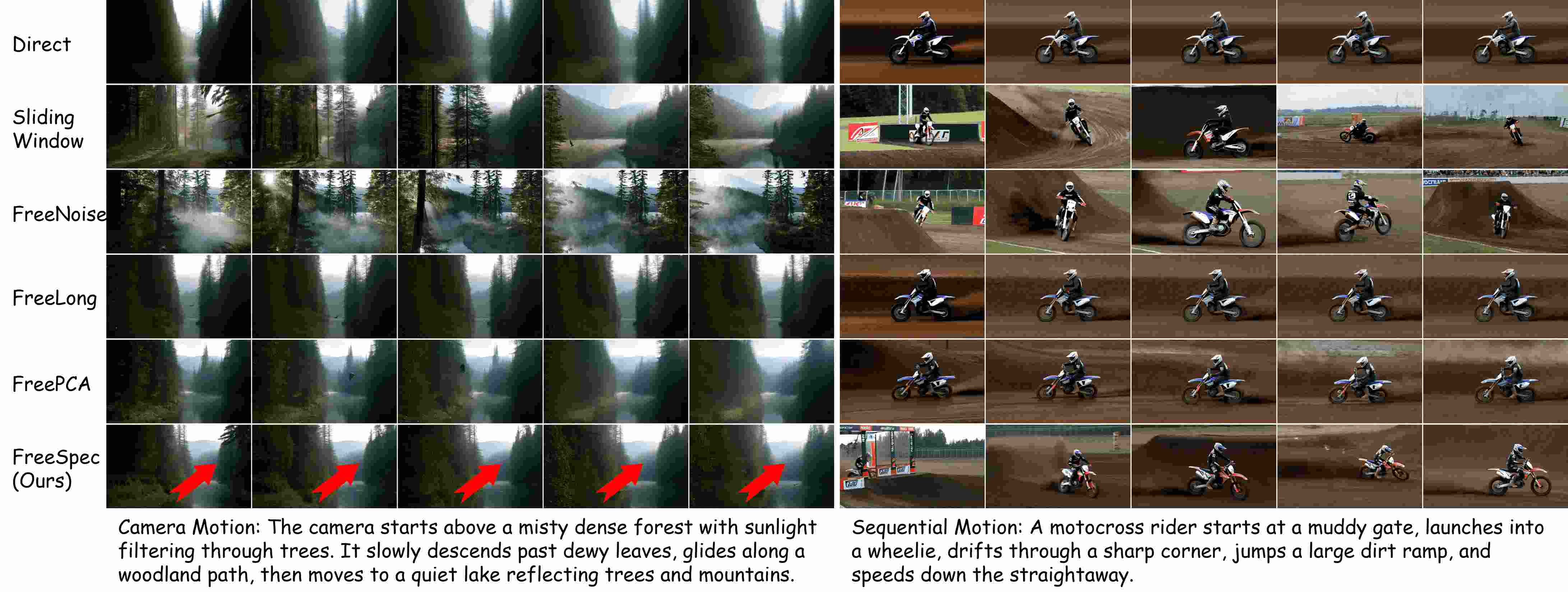}
    \vspace{-1em}
    \caption{
    \textbf{Long-video examples generated on Wan2.1~\cite{wan} with \(4\times\) the native training length.}
    Existing training-free methods preserve stable appearance but may weaken continuous camera trajectories in the forest case and collapse sequential motocross actions into repetitive motion.
    }
    \label{fig:camera_sequential_comparison}
    \vspace{-1em}
\end{figure*}

More closely, FreeLong~\cite{freelong} introduces global-local fusion in the frequency domain, while FreePCA~\cite{freepca} uses Principal Component Analysis (PCA) to select global or local components for stabilizing long-range generation. These methods improve long-range consistency by decomposing appearance consistency and temporal dynamics within each branch using predefined criteria. However, such decomposition can be unreliable because appearance and motion are tightly coupled in video features. For example, FreeLong~\cite{freelong} associates low-frequency signals in the global branch with appearance consistency and high-frequency signals in the local branch with motion details. Nevertheless, slow camera panning and long-range action evolution may also appear in low-frequency components, causing local low-frequency motion cues to be masked by global low-frequency appearance signals. As shown in Fig.~\ref{fig:camera_sequential_comparison}, this limitation weakens continuous viewpoint changes in the forest case and collapses the motocross sequence into repetitive riding patterns. Overall, existing training-free methods remain limited in preserving rich temporal dynamics for long videos.

\begin{figure}[t]
    \centering
    \includegraphics[width=\linewidth]{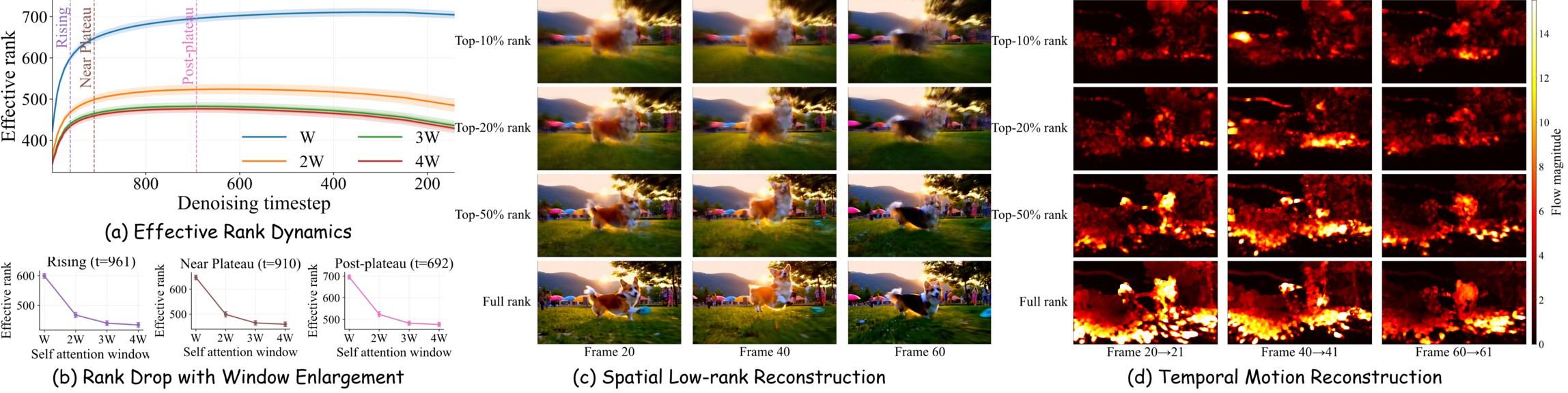}
    \vspace{-1em}
    \caption{
    \textbf{Spectral and qualitative analysis of enlarged self-attention windows.}
    Here, \(W=f\times h\times w\) denotes the native self-attention token length, where \(f\), \(h\), and \(w\) are the temporal length, height, and width of the video latent, respectively.
    (a) shows the effective-rank dynamics across denoising timesteps under different window sizes.
    (b) reports the effective rank at representative timesteps, showing that enlarged windows consistently reduce feature rank.
    (c) and (d) visualize low-rank reconstructions in spatial frames and temporal optical flow, respectively.
    The leading singular components preserve coarse structure but smooth spatial details and weaken temporal variations.
    }
    \label{fig:svd_component_analysis}
    \vspace{-1em}
\end{figure}

To address these challenges, we analyze the distributional properties of global- and local-branch latents. Previous studies~\cite{freelong, freelong++, freepca} show that directly extending the generation length often causes visual blurring and degraded temporal dynamics. Inspired by recent analyses of rank collapse in long-context Transformers~\cite{mind, critical}, we analyze this issue from a singular-spectrum perspective. As shown in Fig.~\ref{fig:svd_component_analysis}, the effective rank increases with long denoising timesteps, but drops rapidly when the attention window is enlarged from the native token length \(W\) to \(2W\), \(3W\), and \(4W\). This indicates that longer contexts induce stronger \textbf{spectral concentration}, where spectral energy is dominated by a few low-rank singular directions. We further reconstruct video features using only the low-rank components to examine this effect. The results show that these components mainly preserve coarse layout and long-range structure, while discarding fine spatial details and motion-rich temporal variations. This explains why enlarged attention windows cause over-smoothed video representations with visual blurring and repetitive patterns: spectral concentration overemphasizes low-rank global structure and suppresses high-rank local variations.

We therefore propose FreeSpec, a training-free singular-spectrum reconstruction framework for long-video generation. FreeSpec mitigates spectral concentration by using the full-window global branch only as low-rank spectral guidance, while preserving the small-window local branch as the high-rank singular basis. Specifically, FreeSpec decomposes the global and local representations with Singular Value Decomposition (SVD) to obtain their singular values and bases. It then performs singular spectrum modulation, injecting global guidance by adjusting singular-value energy. The modulation is controlled by two factors. Timestep-aware modulation emphasizes global structural guidance in early denoising stages and reduces it in later stages for detail and motion refinement. Rank-aware modulation applies stronger global guidance to low-rank singular directions for coarse structure, while preserving higher-rank components for details and motion variations. Finally, the modulated spectrum is reconstructed under the local singular basis, preventing the output from being dominated by the low-rank, over-smoothed global basis.

As illustrated in Appendix Fig.~\ref{fig:fusion_paradigm_comparison}, FreeSpec differs from prior fusion paradigms in whether branch features are partitioned into appearance and motion components. FreeLong~\cite{freelong} fuses low-frequency global components with high-frequency local components, while FreePCA~\cite{freepca} selects cosine-similar PCA components from the global branch and dissimilar ones from the local branch. Both methods use predefined component selection to assign appearance consistency and motion dynamics to different components. In contrast, FreeSpec does not separate appearance and motion. It treats the global branch as spectral guidance and reconstructs the representation under the local singular basis, using low-rank global structure while preserving high-rank local details and temporal dynamics.

Our contributions are summarized as follows:
\begin{itemize}[leftmargin=1.2em, itemsep=0.2em, topsep=0.2em]
    \item We provide a singular-spectrum analysis showing that enlarged self-attention windows induce spectral concentration, which overemphasizes low-rank global structure and suppresses high-rank local variations, causing visual blurring and repetitive patterns.
    
    \item We propose FreeSpec, a training-free spectral reconstruction framework, injects global guidance via timestep- and rank-aware modulation and reconstructs features under local singular basis.

    \item We validate FreeSpec on Wan2.1 and LTX-Video under 4$\times$ length extension, improving temporal smoothness and dynamic degree while maintaining competitive consistency and quality.
\end{itemize}

\section{Related Work}

\subsection{Text-to-Video Generation Models}

Text-to-video generation has advanced rapidly with diffusion-based generative models. Early methods typically adapt pre-trained text-to-image models by adding temporal modules to capture inter-frame dependencies, as shown in VideoLDM~\cite{videoldm}, LaVie~\cite{lavie}, Stable Video Diffusion~\cite{stable_video_diffusion}, AnimateDiff~\cite{animatediff}, and VideoCrafter~\cite{videocrafter1, videocrafter2}. Recent progress has shifted toward diffusion transformers, where CogVideoX~\cite{cogvideox}, Mochi~\cite{mochi}, MovieGen~\cite{moviegen}, HunyuanVideo~\cite{hunyuanvideo}, LTX-Video~\cite{ltx}, and Wan~\cite{wan} demonstrate stronger scalability, motion modeling, and text-video alignment. Despite these advances, most video diffusion models are still trained on short clips and remain limited to several-second generation. Extending them to longer videos is non-trivial because long sequences introduce heavier memory costs and more complex temporal dependencies.

\subsection{Long Video Generation}

Long-video generation has been studied under both training-based and training-free paradigms. Training-based methods~\cite{streamingt2v,vidu, skyreels, magi, rolling_forcing}, such as StreamingT2V~\cite{streamingt2v}, Vidu~\cite{vidu}, and autoregressive or next-clip prediction frameworks~\cite{self_forcing, self_forcing++}, require costly retraining or fine-tuning. In contrast, training-free methods adapt pretrained short-video models only at inference time. Gen-L-Video~\cite{gen_l_video}, FreeNoise~\cite{freenoise}, and FIFO-Diffusion~\cite{fifo} extend generation length using overlapping segments, noise rescheduling, or queue-based denoising, but their local-window design limits long-range consistency. RIFLEx~\cite{riflex} and LongDiff~\cite{longdiff} improve length extrapolation by modifying positional embeddings or temporal position mappings. FreeLOC~\cite{freeloc} further formulates long-video generation as frame-level relative-position and context-length out-of-distribution problems, and introduces layer-adaptive correction. These methods improve length extrapolation and temporal stability, yet offer limited control over the preservation of temporal dynamics.
Closely related to our work, FreeLong~\cite{freelong} and FreeLong++~\cite{freelong++} adopt branch fusion in the frequency domain. It combines low-frequency global components with high-frequency local components to improve long-range stability. FreePCA~\cite{freepca} further replaces frequency-band blending with temporal PCA subspace blending, selecting cosine-similar components from the global branch for consistency and dissimilar components from the local branch for motion preservation. These methods demonstrate the effectiveness of global-local inference for training-free long-video generation. However, their fusion still depends on predefined decomposition criteria, which can be unreliable when structural consistency and temporal dynamics are entangled. In contrast, FreeSpec does not partition appearance and motion within each branch. It treats the global branch as spectral guidance and reconstructs representation under the local singular basis to preserve spatial details and temporal dynamics.

\begin{figure}[t]
    \centering
    \includegraphics[width=\textwidth]{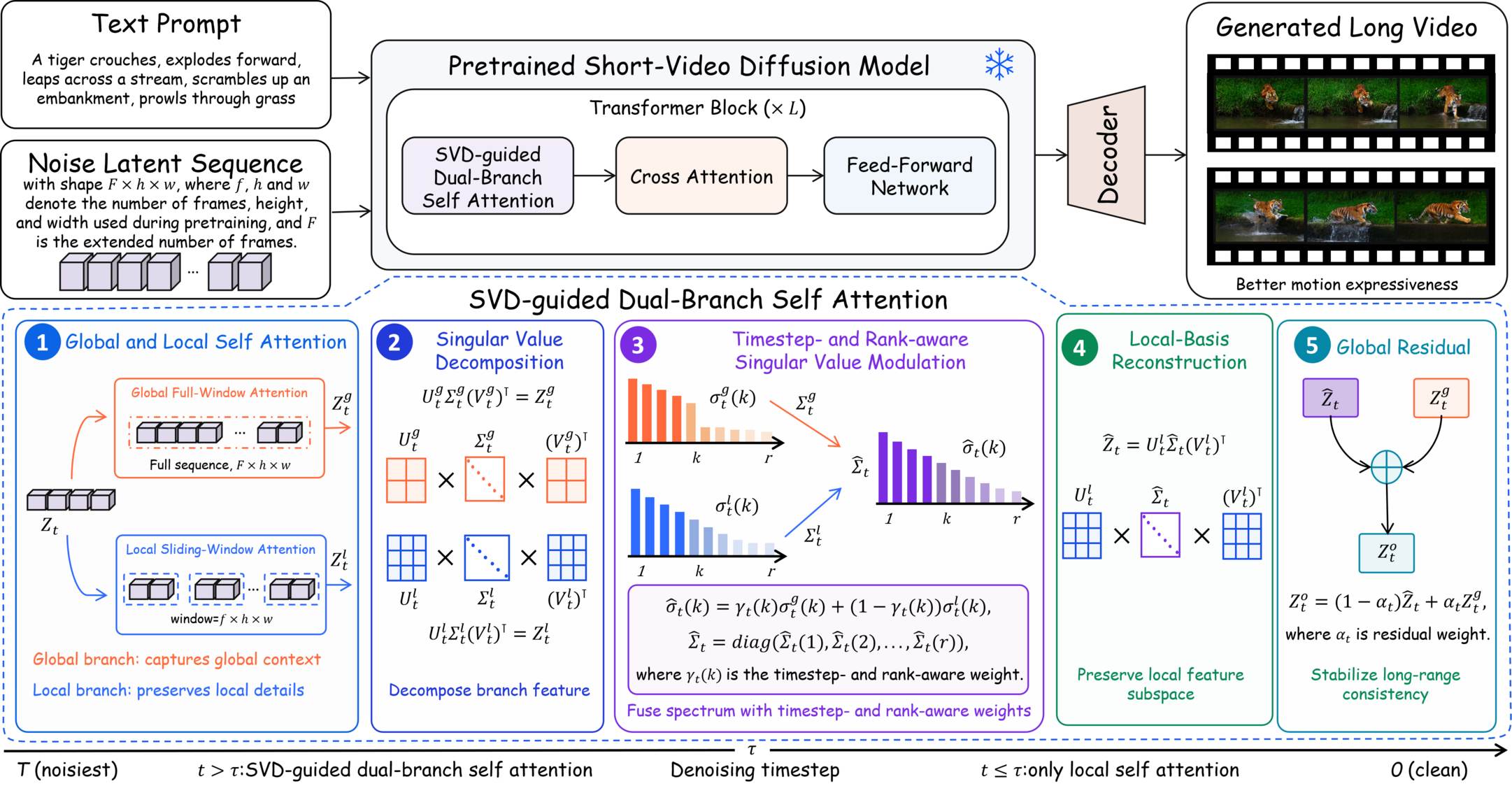}
    \vspace{-1em}
    \caption{
    \textbf{Overview of FreeSpec.}
    FreeSpec extends a frozen short-video diffusion model to long-video generation by replacing self-attention with SVD-guided dual-branch self-attention during inference.
    It combines global full-window guidance and local sliding-window priors through singular-spectrum modulation, followed by local-basis reconstruction and a lightweight global residual.
    }
    \vspace{-1em}
    \label{fig:freesvd_overview}
\end{figure}

\section{Method}

\subsection{Overview}

FreeSpec is motivated by the spectral concentration observed in enlarged attention windows. As shown in Fig.~\ref{fig:freesvd_overview}, FreeSpec keeps the pretrained short-video diffusion model frozen and replaces self-attention layers with SVD-guided dual-branch self-attention during inference. For each replaced layer, FreeSpec constructs a global full-window branch and a local sliding-window branch, decomposes their outputs with SVD, injects global information through timestep- and rank-aware singular-spectrum modulation, and reconstructs the fused representation under the local singular basis. This design allows global guidance to regulate low-rank structural energy while preserving high-rank spatial-temporal variations from the local branch. Following FreePCA~\cite{freepca}, FreeSpec applies dual-branch fusion only in early denoising stages and switches to local-only inference in later steps. This strategy reduces computation with little impact on generation quality in practice. The following sections introduce the main steps of SVD-guided dual-branch self-attention.

\subsection{Dual-Branch Self-Attention} 

Directly applying full-window attention to long videos provides long-range context, but it also induces spectral concentration and causes spatial-temporal over-smoothing. In contrast, sliding-window attention remains close to the pretrained short-video attention regime and better preserves local details and motion-rich variations, but lacks long-range consistency. We therefore adopt a dual-branch strategy that combines a full-window global branch with a sliding-window local branch, using the former for long-range structural guidance and the latter for local detail and temporal dynamics. Let \(F,h,w\) denote the temporal length, height, and width of the long-video latent, respectively, with \(F\gg f\), where \(f\) is the native temporal length used during pretraining. We then define the native token window as \(W=f\times h\times w\).

For query token \(i\), local attention only attends to key-value tokens within its native neighborhood:
\begin{equation}
A_t^l(i,j)=
\begin{cases}
\mathrm{Softmax}\!\left(\frac{\mathbf{Q}_{t,i}\mathbf{K}_{t,j}^{\top}}{\sqrt{d}}\right), & |i-j|\leq W,\\
0, & \mathrm{otherwise}.
\end{cases}
\end{equation}
In contrast, global attention is computed over the entire generated sequence:
\begin{equation}
A_t^g(i,j)=
\mathrm{Softmax}\!\left(\frac{\mathbf{Q}_{t,i}\mathbf{K}_{t,j}^{\top}}{\sqrt{d}}\right).
\end{equation}
The two branch outputs are
\begin{equation}
\mathbf{Z}^{l}_t=A_t^l\mathbf{V}_t,\qquad
\mathbf{Z}^{g}_t=A_t^g\mathbf{V}_t .
\end{equation}
Here, \(\mathbf{Q}_t\), \(\mathbf{K}_t\), and \(\mathbf{V}_t\) are query, key, and value features at timestep \(t\), and \(d\) is the feature dimension. The local branch stays close to the pretrained short-video attention regime and preserves fine details and motion-rich variations, while the global branch provides long-range context for consistency.

\subsection{SVD-guided Spectral Fusion}

\textbf{Singular Value Decomposition.} Our analysis shows that enlarged attention windows induce spectral concentration, where spectral energy is dominated by a few low-rank singular directions. This stabilizes coarse long-range structure but suppresses high-rank spatial-temporal variations. To explicitly operate on this spectrum, FreeSpec decomposes the global and local attention outputs with SVD:
\begin{equation}
\mathbf{U}^{g}_t
\mathbf{\Sigma}^{g}_t
(\mathbf{V}^{g}_t)^\top
=
\mathbf{Z}^{g}_t,
\end{equation}
\begin{equation}
\mathbf{U}^l_t
\mathbf{\Sigma}^l_t
(\mathbf{V}^l_t)^\top
=
\mathbf{Z}^l_t.
\end{equation}
Here, \(\mathbf{U}^{g}_t\) and \(\mathbf{V}^{g}_t\) are the singular-vector matrices of the global output \(\mathbf{Z}^{g}_t\), and \(\mathbf{U}^{l}_t\) and \(\mathbf{V}^{l}_t\) are those of the local output \(\mathbf{Z}^{l}_t\). The diagonal matrices \(\mathbf{\Sigma}^{g}_t\) and \(\mathbf{\Sigma}^{l}_t\) contain the corresponding singular values, which describe the spectral energy distribution of each branch. This decomposition enables singular-value modulation, allowing global structural guidance to be injected selectively.

\textbf{Timestep- and Rank-aware Singular Value Modulation.} Since low-rank singular directions mainly encode coarse structure while high-rank components preserve details and motion variations, and denoising models typically form global structure in early steps and refine details in later steps, global guidance should be introduced selectively rather than uniformly. FreeSpec therefore modulates singular values according to both denoising timestep and rank position.

Given a split timestep \(\tau\), for \(\tau < t < T\), where \(T\) is the maximum timestep, the normalized progress within the fusion stage is
\begin{equation}
p_t=\frac{T-t}{T-\tau}.
\end{equation}
Since the spectral distribution evolves nonlinearly along denoising timesteps, we use a monotonic exponential schedule to gradually shift the fusion from global structural guidance to local detail:
\begin{equation}
w^l_t
=
\frac{1-\exp(-\alpha p_t)}
{1-\exp(-\alpha)},
\qquad
w^{g}_t
=
1-w^l_t,
\end{equation}
where \(\alpha\) controls the transition speed. Early denoising steps use stronger global guidance for structure formation, while later steps rely more on the local branch for detail and motion refinement.

Motivated by the rapid rank reduction under enlarged attention windows, we introduce a rank-aware decay to concentrate global guidance on low-rank singular components and suppress its influence on high-rank components. Let \(r\) be the number of singular values. The global coefficient for the \(k\)-th singular value is
\begin{equation}
\gamma_t(k)
=
w^{g}_t
\exp\left(
-\beta \frac{k}{r}
\right),
\end{equation}
where \(\beta\) controls the decay rate along the spectrum. Let \(\sigma^{g}_t(k)\) and \(\sigma^l_t(k)\) denote the \(k\)-th singular values of the global and local branches. The fused singular value is computed as
\begin{equation}
\hat{\sigma}_t(k)
=
\gamma_t(k)\sigma^{g}_t(k)
+
\left(1-\gamma_t(k)\right)\sigma^l_t(k).
\end{equation}
The fused diagonal matrix is then
\begin{equation}
\hat{\mathbf{\Sigma}}_t
=
\mathrm{diag}
\left(
\hat{\sigma}_t(1),
\hat{\sigma}_t(2),
\dots,
\hat{\sigma}_t(r)
\right).
\end{equation}

\(\gamma_t(k)\) decreases as denoising timestep \(t\) decreases and rank position \(k\) increases. Therefore, the fused singular value \(\hat{\sigma}_t(k)\) shifts toward the local singular value \(\sigma_t^l(k)\) at later denoising steps and higher-rank positions. This design injects global structural guidance mainly into low-rank singular components and early denoising stages. For high-rank components and later timesteps, the fused spectrum is dominated by the local branch, preserving spatial details and temporal variations.

\textbf{Local-Basis Reconstruction.} After singular-spectrum modulation, the remaining question is which singular basis should reconstruct the fused feature. Since the global branch is more affected by spectral concentration, using its basis may project the output toward a low-rank and over-smoothed representation. FreeSpec therefore reconstructs the fused feature under the local singular basis:
\begin{equation}
\hat{\mathbf{Z}}_t
=
\mathbf{U}^l_t
\hat{\mathbf{\Sigma}}_t
(\mathbf{V}^l_t)^\top.
\end{equation}
This local-basis reconstruction preserves high-rank spatial-temporal variations from the local branch, while the modulated singular values still carry controlled global structural guidance.

\textbf{Global Residual.} Local-basis reconstruction preserves high-rank details and motion dynamics, but the reconstructed feature may still lose part of the full-window structural signal. FreeSpec therefore adds a lightweight global residual:
\begin{equation}
a_t
=
a_0+a_1 w^{g}_t,
\end{equation}
\begin{equation}
\mathbf{Z}^{o}_t
=
(1-a_t)\hat{\mathbf{Z}}_t
+
a_t\mathbf{Z}^{g}_t.
\end{equation}
Here, \(a_0\) and \(a_1\) control the residual strength, and \(a_t\) is the timestep-dependent residual weight. The final output \(\mathbf{Z}^{o}_t\) keeps the local-basis reconstruction as the main representation while retaining a small amount of global full-window information for long-range stabilization.

Similar to FreePCA~\cite{freepca}, FreeSpec performs SVD-guided dual-branch self-attention only for \(\tau < t < T\), corresponding to the early structure-forming denoising stage. For \(t \leq \tau\), it uses the local branch only. This schedule improves inference efficiency.

\section{Experiments}

This section presents the setup, main results, and ablations. More details, hyperparameter analysis, qualitative results, and a demo website are provided in the supplementary material.

\subsection{Experimental Setup}

\textbf{Datasets.}
Following recent training-free long video generation methods~\cite{freelong, freelong++, freepca}, we evaluate our method on long text-to-video generation using 100 enhanced prompts from VBench-Long~\cite{vbench++}. 

\textbf{Compared Methods.}
We compare FreeSpec with representative training-free long video generation baselines.
\textbf{Direct} directly applies full self-attention to the extended video sequence without any adaptation.
\textbf{Sliding Window} restricts self-attention to a local window whose length is the same as the training length \(W\).
\textbf{FreeNoise}~\cite{freenoise} extends short-video diffusion models through noise rescheduling and window-based attention fusion.
\textbf{FreeLong}~\cite{freelong} adopts a global-local inference paradigm and fuses the low-frequency components of the global branch with the high-frequency components of the local branch in the frequency domain.
\textbf{FreePCA}~\cite{freepca} decomposes global and local features into a shared principal subspace and progressively injects global consistency components into local features according to component-level similarity.

\textbf{Evaluation Metrics.}
We adopt VBench-based metrics~\cite{vbench,vbench++} to evaluate video consistency, temporal quality, and perceptual quality. For consistency, we use \textbf{Subject Consistency}, which measures cross-frame object appearance similarity using DINO features~\cite{dino}, and \textbf{Background Consistency}, which measures scene-level background stability using CLIP features~\cite{clip}. For temporal quality, we report \textbf{Motion Smoothness}, which uses motion priors from a frame interpolation model to assess whether generated motion evolves smoothly~\cite{amt}, and \textbf{Dynamic Degree}, which quantifies motion intensity via optical flow magnitude~\cite{raft}. For perceptual quality, we report \textbf{Aesthetic Quality}, which measures artistic merit using the LAION aesthetic predictor~\cite{aestheticpredictor}, and \textbf{Imaging Quality}, which evaluates frame-level visual quality using an image quality predictor~\cite{musiq}.

\begin{table}[t]
\centering
\caption{\textbf{Quantitative comparison on Wan2.1-1.3B ($4 \times$).} The best result is highlighted in \textbf{bold}, and the second-best result is \underline{underlined}. Inference time denotes the runtime on an A100 GPU.}
\label{tab:main_results}
\resizebox{\linewidth}{!}{
\begin{tabular}{lccccccc}
\toprule
\textbf{Method} 
& \textbf{Subj.-Con.} 
& \textbf{Bkgd.-Con.} 
& \textbf{Mot.-Smth.} 
& \textbf{Dyn.-Deg.} 
& \textbf{Aes.-Qual.} 
& \textbf{Img.-Qual.}
& \textbf{Infer. Time} \\
\midrule
\textbf{Direct} 
& \textbf{0.9865} 
& \textbf{0.9752} 
& 0.9876 
& 0.0468 
& 0.5858 
& 0.6274
& $\sim$38 mins \\
\textbf{Sliding Window} 
& 0.9791 
& 0.9677 
& \underline{0.9887} 
& \textbf{0.3955} 
& 0.6018 
& \underline{0.6424}
& $\sim$25 mins \\
\textbf{FreeNoise} 
& 0.9777 
& 0.9667 
& 0.9847 
& 0.2395 
& \underline{0.6074} 
& \textbf{0.6612}
& $\sim$25 mins \\
\textbf{FreeLong} 
& 0.9849 
& 0.9731 
& 0.9856 
& 0.2514 
& 0.6057 
& 0.6374
& $\sim$61 mins \\
\textbf{FreePCA} 
& 0.9857 
& 0.9734 
& 0.9853 
& 0.2177 
& 0.6005 
& 0.6372
& $\sim$55 mins \\
\textbf{FreeSpec (Ours)} 
& \underline{0.9863} 
& \underline{0.9745} 
& \textbf{0.9902} 
& \underline{0.3200} 
& \textbf{0.6104} 
& 0.6415
& $\sim$56 mins \\
\bottomrule
\end{tabular}
}
\end{table}

\begin{table}[t]
\centering
\caption{\textbf{Quantitative comparison on LTX-Video-2B-dev ($4 \times$).} The best result is highlighted in \textbf{bold}, and the second-best result is \underline{underlined}. Inference time denotes the runtime on an A100 GPU.}
\label{tab:main_results_ltx}
\resizebox{\linewidth}{!}{
\begin{tabular}{lccccccc}
\toprule
\textbf{Method} 
& \textbf{Subj.-Con.} 
& \textbf{Bkgd.-Con.} 
& \textbf{Mot.-Smth.} 
& \textbf{Dyn.-Deg.} 
& \textbf{Aes.-Qual.} 
& \textbf{Img.-Qual.}
& \textbf{Infer. Time} \\
\midrule
\textbf{Direct} 
& \textbf{0.9726} 
& \textbf{0.9682} 
& 0.9800 
& 0.1933 
& 0.5448 
& 0.6468
& $\sim$13 mins \\
\textbf{Sliding Window} 
& 0.9559 
& 0.9620 
& 0.9795 
& \underline{0.2311} 
& 0.5442 
& 0.6282
& $\sim$12 mins \\
\textbf{FreeNoise} 
& 0.9618 
& 0.9614 
& \underline{0.9822} 
& 0.0556 
& 0.5230 
& \textbf{0.6528}
& $\sim$12 mins \\
\textbf{FreeLong} 
& \underline{0.9699} 
& 0.9640 
& 0.9804 
& 0.0311 
& 0.5430 
& \underline{0.6478}
& $\sim$22 mins \\
\textbf{FreePCA} 
& 0.9610 
& 0.9637 
& 0.9796 
& 0.2111 
& \underline{0.5453} 
& 0.6332
& $\sim$21 mins \\
\textbf{FreeSpec (Ours)} 
& 0.9622 
& \underline{0.9677} 
& \textbf{0.9892} 
& \textbf{0.2478} 
& \textbf{0.5464} 
& 0.6382
& $\sim$22 mins \\
\bottomrule
\end{tabular}
}
\vspace{-1em}
\end{table}

\begin{figure*}[t]
    \centering
    \hspace*{-0.05\linewidth}
    \includegraphics[width=0.95\linewidth]{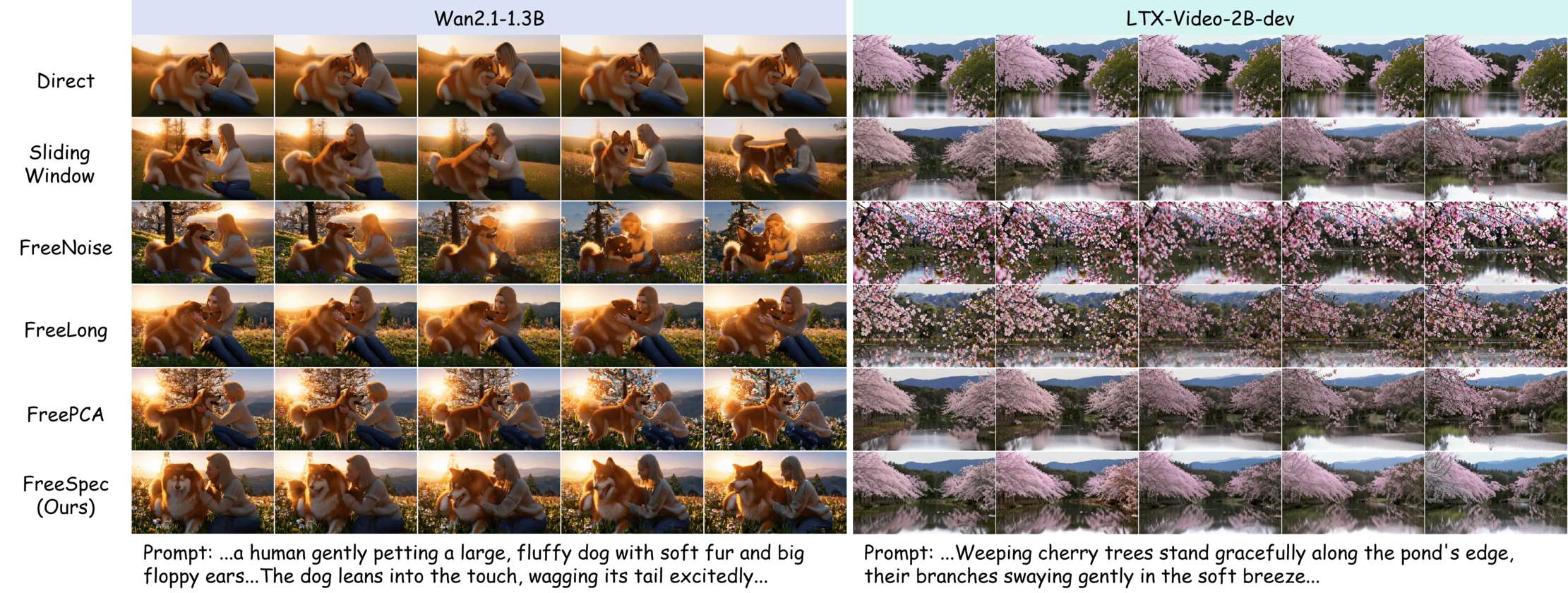}
    \caption{\textbf{Qualitative comparison under 4$\times$ length extension on Wan2.1 and LTX-Video.}}
    \vspace{-1em}
    \label{fig:qualitative_comparison}
\end{figure*}

\subsection{Main Results}

We evaluate FreeSpec on two public text-to-video backbones, Wan2.1-1.3B~\cite{wan} and LTX-Video-2B-dev~\cite{ltx}. For each backbone, we generate videos with \(4\times\) the native training length, i.e., 321 frames for Wan2.1-1.3B and 481 frames for LTX-Video-2B-dev.

\textbf{Quantitative comparison.}
Tables~\ref{tab:main_results} and~\ref{tab:main_results_ltx} report the quantitative results. Direct full attention achieves strong subject and background consistency but yields low dynamic degree, confirming that enlarged attention windows tend to over-smooth temporal evolution. Sliding Window preserves motion within the native temporal range, but weakens long-range consistency. FreeNoise improves visual quality on some metrics, yet its consistency and motion scores remain limited, indicating that noise rescheduling alone is insufficient to preserve spatial consistency and temporal dynamics. FreeLong and FreePCA obtain competitive consistency by using global-local fusion, but their frequency- or PCA-based decomposition can still suppress useful motion cues, leading to weaker dynamic degree. In contrast, FreeSpec achieves the best motion smoothness on both backbones and improves dynamic degree over most global-local baselines, while maintaining competitive consistency and perceptual quality. These results show that FreeSpec uses global information as low-rank structural guidance and preserves high-rank local variations through local-basis reconstruction.

\textbf{Qualitative comparison.}
Fig.~\ref{fig:qualitative_comparison} shows consistent qualitative trends. Direct full attention produces stable but nearly static videos with visual blurring, while Sliding Window introduces more motion at the cost of appearance and layout stability. FreeNoise often improves frame-level quality, but action progression remains weak. FreeLong and FreePCA better stabilize global structure, yet their decomposition-based fusion may still produce conservative motion or incomplete temporal evolution. By contrast, FreeSpec preserves subject appearance and long-range layout while producing richer temporal changes, such as smoother human-dog interactions on Wan2.1-1.3B, and maintaining stronger long-range consistency in the relatively static landscape sequence on LTX-Video-2B-dev. These results further demonstrate that FreeSpec mitigates over-smoothing while preserving motion-rich temporal dynamics in dynamic scenes and long-range consistency in relatively static scenes.

\begin{table*}[t]
\centering
\caption{\textbf{Ablation results on the first 50 prompts of the evaluation set using Wan2.1-1.3B.} 
\textbf{Bold} and \underline{underline} indicate the best and second-best results, respectively.}
\label{tab:ablation}
\setlength{\tabcolsep}{4.2pt}
\renewcommand{\arraystretch}{1.12}
\resizebox{\textwidth}{!}{
\begin{tabular}{llcccccc}
\toprule
\textbf{Component} & \textbf{Variant}
& \textbf{Subj.-Con. }
& \textbf{Bkgd.-Con. }
& \textbf{Mot.-Smth. }
& \textbf{Dyn.-Deg. }
& \textbf{Aes.-Qual.} 
& \textbf{Img.-Qual.} \\
\midrule

\textbf{Global basis}
& GB
& \textbf{0.9931}
& \textbf{0.9817}
& \textbf{0.9904}
& 0.2182
& \textbf{0.6259}
& \textbf{0.6749} \\

\midrule
\textbf{Local basis}
& LB + fixed fusion
& 0.9821
& 0.9696
& 0.9895
& 0.3782
& 0.6175
& 0.6471 \\

\midrule
\multirow{3}{*}{\textbf{Weighted modulation}}
& LB + RA
& 0.9820
& 0.9696
& 0.9897
& 0.3455
& 0.6078
& 0.6212 \\
& LB + TA
& 0.9810
& 0.9687
& 0.9895
& 0.3800
& 0.6196
& 0.6458 \\
& LB + RA + TA
& 0.9809
& 0.9688
& 0.9894
& 0.3836
& 0.6189
& \underline{0.6494} \\

\midrule
\multirow{2}{*}{\textbf{Lightweight residual}}
& LB + RA + TA + F-GR
& 0.9857
& 0.9737
& 0.9898
& \underline{0.3891}
& 0.6191
& 0.6432 \\
& LB + RA + TA + T-GR
& \underline{0.9860}
& \underline{0.9742}
& \underline{0.9899}
& \textbf{0.3891}
& \underline{0.6207}
& 0.6462 \\

\bottomrule
\end{tabular}
}
\vspace{-1.5em}
\end{table*}

\begin{figure*}[t]
    \centering
    \includegraphics[width=\textwidth]{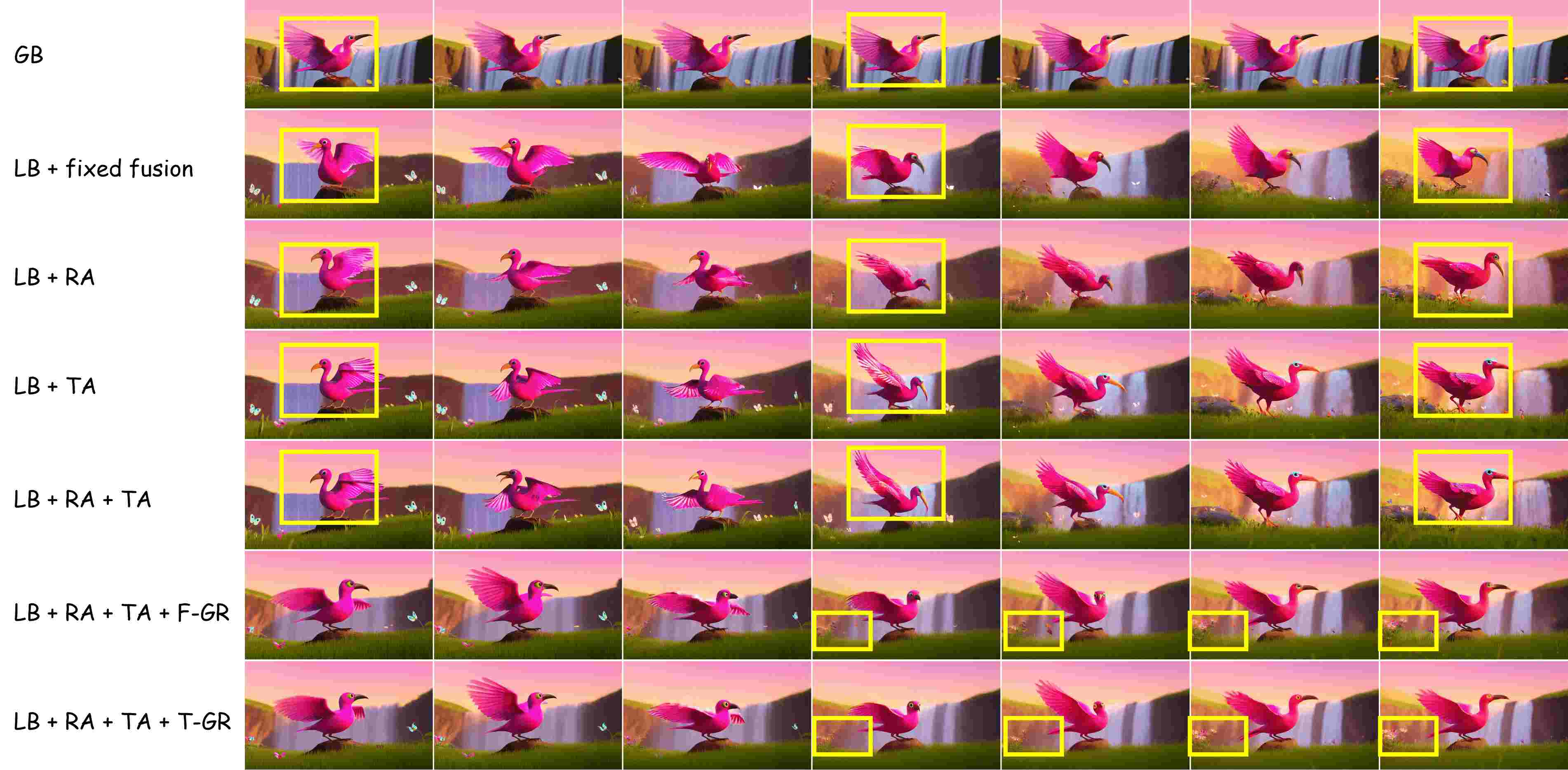}
    \caption{
    Qualitative ablation results of FreeSpec on Wan2.1. Yellow boxes highlight representative regions with subject inconsistency, background drift, local artifacts, or weakened temporal evolution.
    }
    \vspace{-1em}
    \label{fig:qualitative_ablation}
\end{figure*}

\subsection{Ablation Study}

We conduct ablations on the first 50 evaluation prompts to examine each component of FreeSpec. \textbf{GB} and \textbf{LB} denote reconstruction with the global and local singular bases. \textbf{TA} and \textbf{RA} denote timestep- and rank-aware singular-spectrum modulation, respectively. \textbf{F-GR} and \textbf{T-GR} denote fixed and timestep-aware global residuals.

\textbf{Quantitative ablation.}
Table~\ref{tab:ablation} reports the results. The global-basis variant obtains the best consistency and perceptual quality, but its dynamic degree is much lower than all local-basis variants. This confirms that the global singular basis favors low-rank structural stability and tends to suppress high-rank temporal variations. In contrast, local-basis reconstruction substantially improves dynamic degree, showing that the local singular basis is better suited for preserving motion-rich short-video priors, although global guidance is still needed for stronger consistency.
For singular-spectrum modulation, TA is more effective than RA alone, since it adapts global guidance across denoising stages and shifts the fusion from early structural guidance to later local refinement. Combining TA and RA further improves the motion-stability trade-off by restricting global guidance mainly to low-rank singular components while preserving high-rank local variations. Finally, adding global residual improves consistency over pure local-basis reconstruction. Compared with F-GR, T-GR achieves a better balance by applying stronger residual anchoring when global structure is needed and reducing it during later refinement.

\textbf{Qualitative ablation.}
Fig.~\ref{fig:qualitative_ablation} provides consistent qualitative evidence. GB produces stable appearance and background layout but weakens wing motion, indicating over-reliance on low-rank global structure. LB restores clearer pose changes and stronger wing dynamics, but introduces local distortions and background inconsistency. TA reduces these artifacts by controlling global guidance over denoising timesteps, while RA further protects high-rank motion details from excessive global influence. With global residual, especially T-GR, the sequence maintains more consistent subject appearance and background structure while preserving visible wing motion. These results verify that FreeSpec benefits from using the local singular basis as the main reconstruction space, with global information introduced only through controlled spectral guidance and lightweight residual.

\begin{figure}[t]
    \centering
    \includegraphics[width=\linewidth]{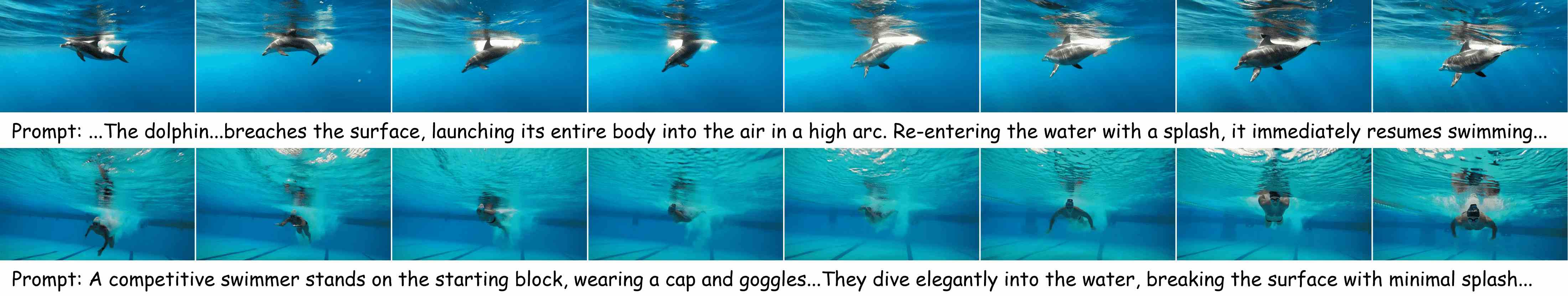}
    \caption{
    \textbf{Failure cases.} FreeSpec preserves temporal dynamics but fails to infer implicit scene transitions, such as underwater-to-air and diving-platform-to-pool transitions.
    }
    \vspace{-1em}
    \label{fig:failure_cases}
\end{figure}

\section{Failure Cases and Limitations}
\label{sec:failure_cases}

Although FreeSpec improves long-range temporal dynamics, it still inherits the semantic limitations of the pretrained base model. As shown in Fig.~\ref{fig:failure_cases}, the first case requires a transition from underwater to air, but the generated dolphin remains underwater throughout the sequence, showing only local pose and motion changes. In the second case, the action should progress from a diving platform to a swimming pool, yet the video stays directly in the underwater pool scene and omits the platform stage. These examples show that FreeSpec can preserve temporal dynamics, such as swimming trajectories and body poses, but cannot reliably infer implicit scene transitions in sequential actions. This limitation mainly stems from the restricted semantic reasoning ability of the base model, which also fails to identify such hidden scene changes. As a training-free feature modulation method, FreeSpec adjusts intermediate temporal representations but does not introduce event-level planning or explicit scene-transition control. Future work may incorporate stronger language understanding or scene-level control signals to handle long videos with complex semantic progression.

\vspace{-1em}
\section{Conclusion}

In this paper, we presented FreeSpec, a training-free spectral reconstruction framework for long-video generation. We showed that enlarged self-attention windows cause spectral concentration, where feature energy is dominated by a few low-rank singular directions. This preserves coarse structure but suppresses high-rank details and temporal variations, resulting in spatial smoothing and weakened dynamics. To address this, FreeSpec uses the global branch only as low-rank spectral guidance and reconstructs features under the local singular basis. Timestep- and rank-aware modulation further controls global guidance across denoising stages and singular directions. Experiments on Wan2.1 and LTX-Video demonstrate that FreeSpec achieves a better balance between long-range consistency, motion dynamics, and visual quality. However, as a training-free method, it is still limited by the semantic reasoning ability of the frozen base model, especially for implicit scene transitions in sequential actions.

\clearpage

\bibliographystyle{unsrtnat}
\bibliography{references}


\clearpage
\setcounter{page}{1}
\setcounter{table}{0}
\setcounter{figure}{0}
\appendix

\renewcommand{\thetable}{A\arabic{table}}
\renewcommand{\thefigure}{A\arabic{figure}}

\section{Supplementary Overview}

This supplementary material provides additional analysis, implementation details, and qualitative results for FreeSpec. It includes:

\begin{itemize}[leftmargin=1.2em, itemsep=0.2em, topsep=0.2em]
    \item Sec.~\ref{app:observation} expands the spectral observation in the main paper with effective-rank analysis and low-rank reconstruction results.
    \item Sec.~\ref{app:implementation} reports implementation details and default hyperparameters.
    \item Sec.~\ref{app:method_comparison} provides a visual comparison between FreeSpec and representative training-free fusion paradigms.
    \item Sec.~\ref{app:experiment} includes hyperparameter analysis and additional qualitative comparisons on complex-motion and VBench-Long cases.
\end{itemize}

\section{Observation and Analysis}
\label{app:observation}

This section provides a more detailed spectral analysis of enlarged self-attention windows. 
We examine how temporal context extension affects the singular spectrum of intermediate video diffusion features, and further visualize the role of different spectral components through spatial-frame and optical-flow reconstructions.

Motivated by recent spectral analyses of long-context transformers~\cite{mind,critical}, we examine how context extension affects intermediate video representations in diffusion-based video generation. Prior studies suggest that enlarging the attention context may lead to more uniform attention scores, excessive token aggregation, and rank collapse along the context-length dimension. Since training-free long-video generation often extends pretrained short-video models by enlarging attention windows during inference, we investigate whether similar spectral degeneration also appears in video diffusion features.

Specifically, we analyze the singular spectrum of intermediate representations under different temporal self-attention windows, including \(W\), \(2W\), \(3W\), and \(4W\). Here, \(W=f\times h\times w\) denotes the native self-attention token length, where \(f\), \(h\), and \(w\) denote the temporal length, height, and width used during pretraining, respectively. Enlarging the window from \(W\) to \(4W\) therefore extends the temporal receptive field from the native pretrained context to a four-times longer token context. We compute the effective rank over denoising timesteps on 20 prompts and report the mean curve with standard error. To further interpret the role of different spectral components, we reconstruct video representations using different rank ratios and visualize both spatial frames and temporal optical flow.

\paragraph{Effective rank follows temporal growth but decays with window enlargement.}
We first study how denoising progress and temporal inference window jointly affect the rank structure of video representations. Given singular values \(\{\sigma_i\}_{i=1}^{n}\), we normalize them as \(p_i=\sigma_i/\sum_j\sigma_j\), and compute the effective rank as \(r_{\mathrm{eff}}=\exp(-\sum_i p_i\log p_i)\). A larger effective rank indicates that representation energy is distributed across more singular directions, whereas a smaller value indicates stronger concentration on a few dominant components.

As shown in Fig.~\ref{fig:svd_component_analysis}(a), for each fixed self-attention window, the effective rank increases rapidly during early denoising and then gradually approaches a plateau, forming a growth pattern over denoising timesteps. This suggests that video representations progressively activate more singular directions as generation evolves from coarse structure formation to detailed content refinement. However, this temporal growth is strongly modulated by the attention window size. Enlarging the window from \(W\) to \(2W\), \(3W\), and \(4W\) consistently lowers the effective-rank curve, indicating that longer temporal context strengthens token aggregation and concentrates representation energy into fewer dominant singular components.

Fig.~\ref{fig:svd_component_analysis}(b) further shows this window-dependent effect at representative denoising stages, including the rising, near-plateau, and post-plateau phases. Across all stages, the effective rank drops sharply when the window is enlarged from \(W\) to \(2W\), and then decreases more slowly from \(2W\) to \(4W\), exhibiting an exponential-like decay with respect to window size. This consistent rank decay indicates that temporal context extension does not only provide longer-range dependencies, but also drives intermediate video representations toward a more concentrated low-rank structure.

\paragraph{Low-rank components preserve coarse structure but suppress details and motion.}
We then analyze the semantic role of different spectral components by reconstructing video representations with different rank ratios. As shown in Fig.~\ref{fig:svd_component_analysis}(c), the full-rank reconstruction preserves clear spatial appearance, including object boundaries, local textures, and background details. When only the top-\(50\%\) singular components are retained, the global layout and main subject structure remain recognizable, but local details become visibly smoother. This degradation becomes more pronounced in the top-\(20\%\) and top-\(10\%\) reconstructions, where frames preserve only coarse appearance and lose fine-grained texture.

A similar tendency can be observed in the temporal optical-flow visualization in Fig.~\ref{fig:svd_component_analysis}(d). Full-rank reconstruction produces stronger and more structured flow responses around moving regions, while low-rank reconstructions yield weaker and less detailed flow maps, indicating that motion-rich temporal variations are largely suppressed. These observations suggest that low-rank singular components mainly encode coarse layout and long-range structure, whereas higher-rank components are important for local texture, sharp boundaries, and expressive motion dynamics.

Overall, the analysis reveals a dual effect of temporal context extension. Along denoising timesteps, video representations naturally evolve toward richer rank structures, but enlarged self-attention windows counteract this process by inducing rank decay and stronger spectral concentration. While this low-rank bias may benefit long-range structural consistency, excessive concentration suppresses high-rank components essential for visual fidelity and motion richness. This motivates our spectral reconstruction strategy: instead of directly replacing local representations with global-window outputs, we inject global structural guidance through controlled singular-spectrum modulation while preserving local high-rank components for fine spatial details and dynamic temporal variations.

\section{Implementation Details}
\label{app:implementation}

This section summarizes the default implementation settings used in our experiments. Unless otherwise specified, these hyperparameters are fixed across backbones and evaluation cases.

Unless otherwise specified, the main hyperparameters of FreeSpec are fixed across experiments: the timestep threshold is set to \(\tau=0.9\) with \(T=1.0\), the timestep transition coefficient is set to \(\alpha=5.0\), the rank-aware decay factor is set to \(\beta=5.0\), and the global residual weight follows a timestep-aware schedule parameterized by \(a_0=0.15\) and \(a_1=0.2\).

\begin{figure}[t]
    \centering
    \includegraphics[width=0.95\linewidth]{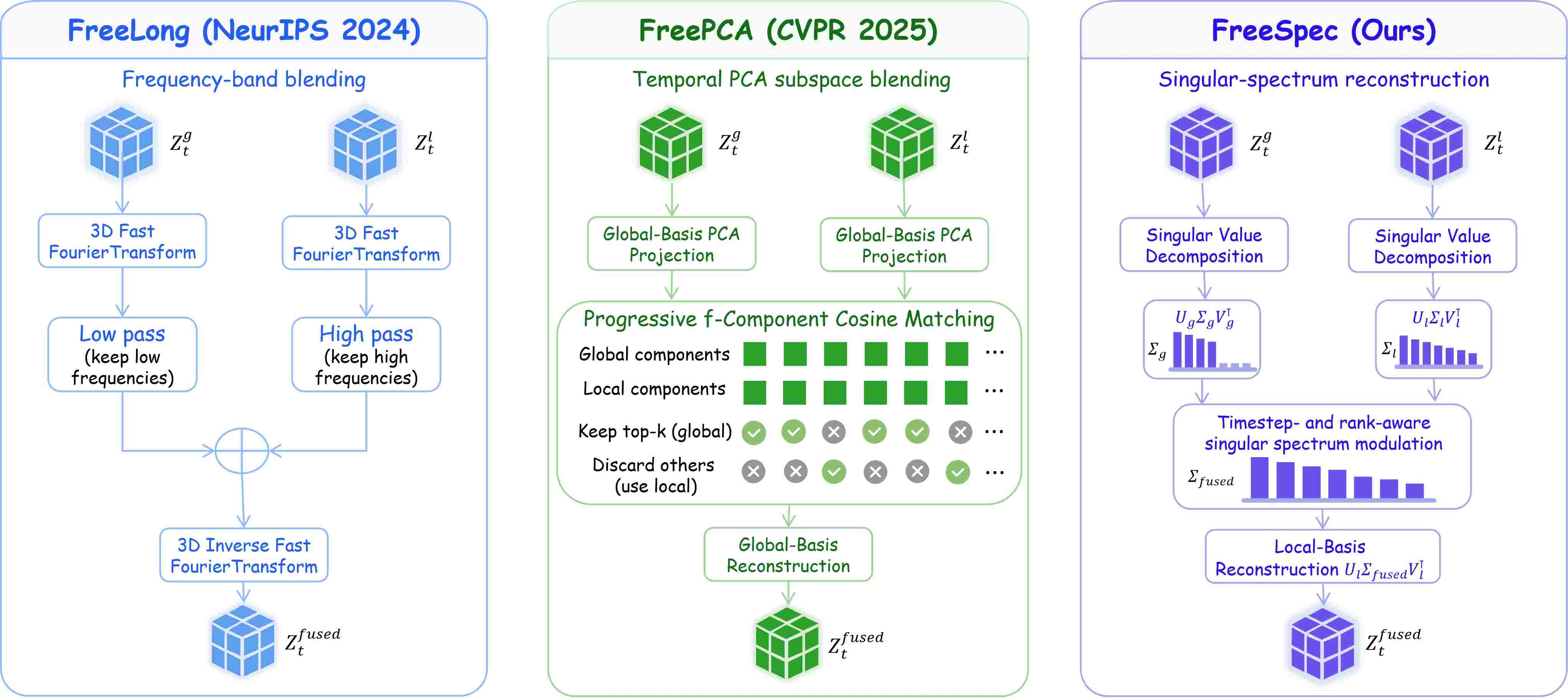}
    \caption{
    Visual comparison of training-free global-local fusion paradigms.
    FreeLong uses frequency-band fusion, FreePCA uses PCA-component selection, and FreeSpec performs singular-spectrum modulation followed by local-basis reconstruction.
    }
    \label{fig:fusion_paradigm_comparison}
\end{figure}

\section{Method Comparison}
\label{app:method_comparison}

This section visualizes the difference between FreeSpec and representative training-free global-local fusion methods. The comparison highlights whether branch features are explicitly partitioned into appearance and motion components.

\section{Additional Experiments}
\label{app:experiment}

This section provides additional experimental results beyond the main paper. We first analyze the sensitivity of FreeSpec to key hyperparameters, and then present more qualitative comparisons on complex-motion and VBench-Long prompts.

\begin{figure}[t]
    \centering
    \includegraphics[width=\linewidth]{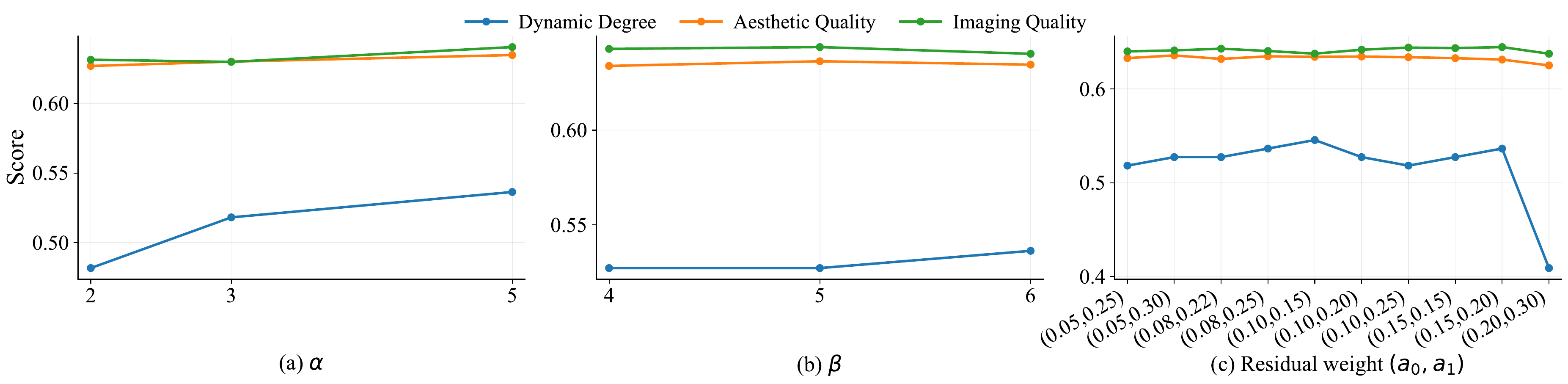}
    \caption{
    Hyperparameter sensitivity of FreeSpec.
    We vary \(\alpha\), \(\beta\), and \(a_t=a_0+a_1w_t^g\), and report dynamic degree, aesthetic quality, and imaging quality.
    }
    \label{fig:hyperparameter_analysis}
\end{figure}

\subsection{Hyperparameter Analysis}

We analyze the sensitivity of FreeSpec to three key hyperparameters: the timestep transition coefficient \(\alpha\), the rank-decay coefficient \(\beta\), and the residual weight \(a_t=a_0+a_1 w_t^g\). As shown in Fig.~\ref{fig:hyperparameter_analysis}, we focus on Dynamic Degree, Aesthetic Quality, and Imaging Quality, which reflect motion richness and perceptual quality.

For \(\alpha\), increasing it from \(2\) to \(5\) consistently improves dynamic degree, with mild gains in aesthetic and imaging quality. This suggests that a sharper transition from global guidance to local dynamics better preserves motion richness in later denoising stages. Although \(\alpha=5\) slightly reduces consistency compared with \(\alpha=2\) and \(\alpha=3\), it achieves the best overall trade-off among motion dynamics and perceptual quality, and is therefore used as the default setting.

For \(\beta\), performance remains stable within the tested range. Increasing \(\beta\) from \(4\) to \(6\) slightly improves dynamic degree, while aesthetic and imaging quality change only marginally. This indicates that FreeSpec is not sensitive to the rank-decay coefficient, and moderate rank-aware modulation is sufficient to suppress global influence on high-rank components.

For the residual weight, the trade-off is more evident. Moderate settings, such as \((0.08,0.25)\), \((0.10,0.15)\), and \((0.15,0.20)\), maintain competitive perceptual quality and motion dynamics. In contrast, an overly strong residual, such as \((0.20,0.30)\), sharply reduces dynamic degree and also degrades aesthetic and imaging quality. This confirms that excessive global residual guidance can overwrite the local branch and weaken temporal dynamics.

\begin{figure*}[t]
    \centering
    \hspace*{-0.05\linewidth}
    \includegraphics[width=0.95\linewidth]{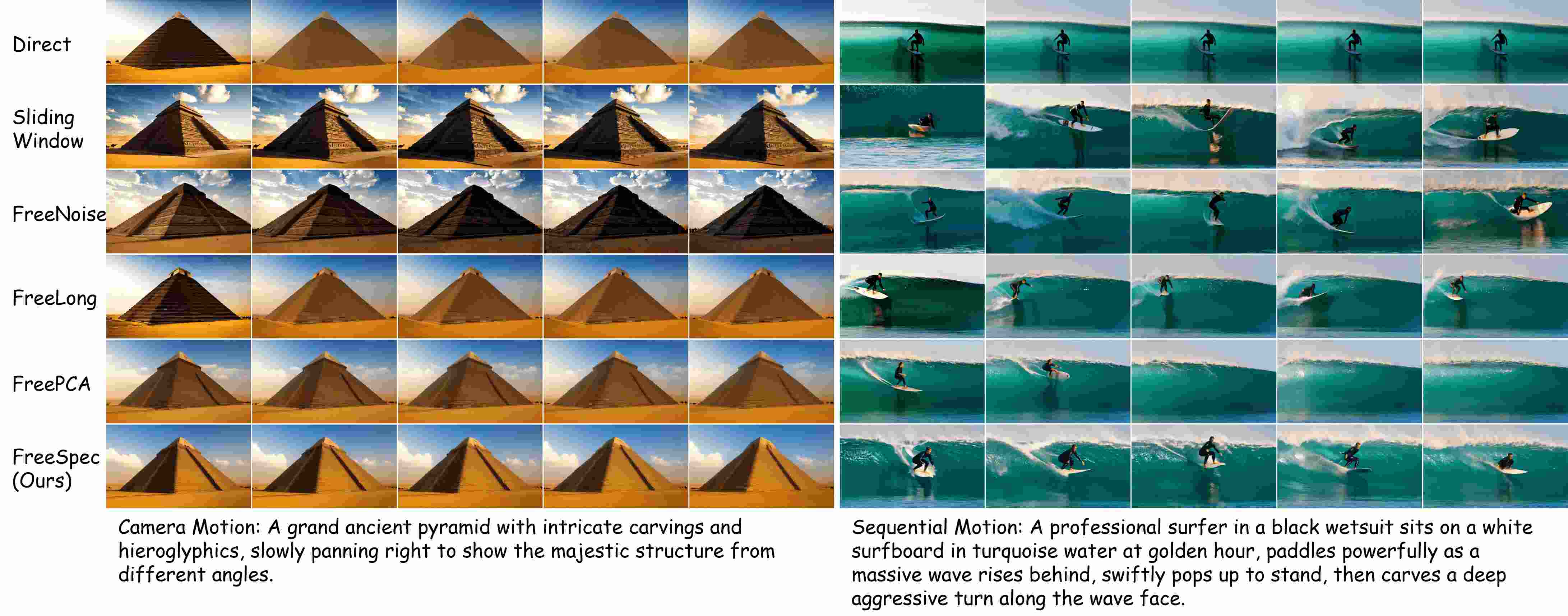}
    \caption{
    Additional complex-motion examples on Wan2.1~\cite{wan}.
    The results compare camera-motion and sequential-motion generation under \(4\times\) the native training length.
    }
    \label{fig:camera_sequential_comparison_2}
\end{figure*}

\subsection{Additional Complex-Motion Results}

This section provides additional qualitative examples on camera-motion and sequential-motion prompts. These cases are designed to evaluate whether a method can preserve complex temporal dynamics when extending Wan2.1~\cite{wan} to \(4\times\) its native training length.

\begin{figure*}[t]
    \centering
    \hspace*{-0.05\linewidth}
    \includegraphics[width=0.95\linewidth]{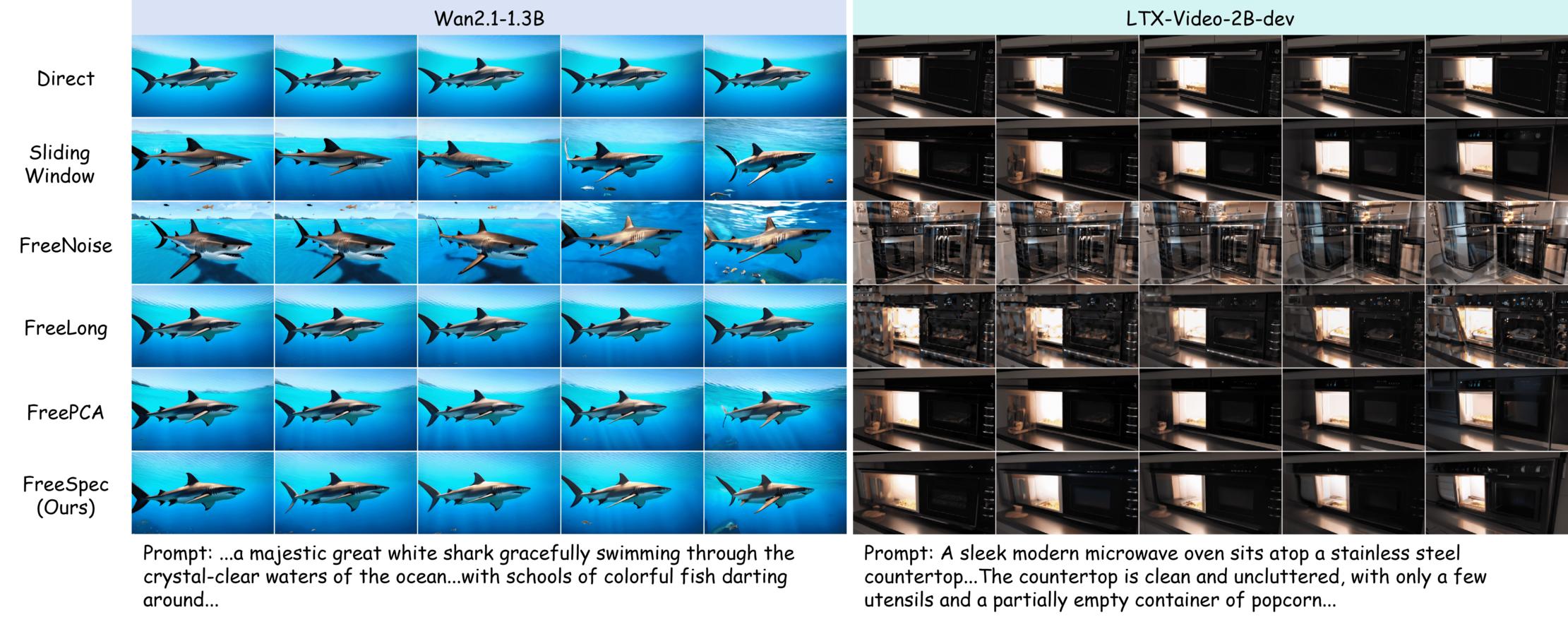}
    \caption{
    Additional qualitative comparison for long-video generation.
    }
    \label{fig:qualitative_comparison_2}
\end{figure*}

\subsection{Additional Qualitative Comparisons}

This section provides more qualitative comparisons on different prompts and backbones. The results complement the main paper by showing the behavior of each method under diverse long-video generation scenarios.

\begin{figure*}[t]
    \centering
    \foreach \idx in {3,11,45,96}{
        \includegraphics[width=\linewidth]{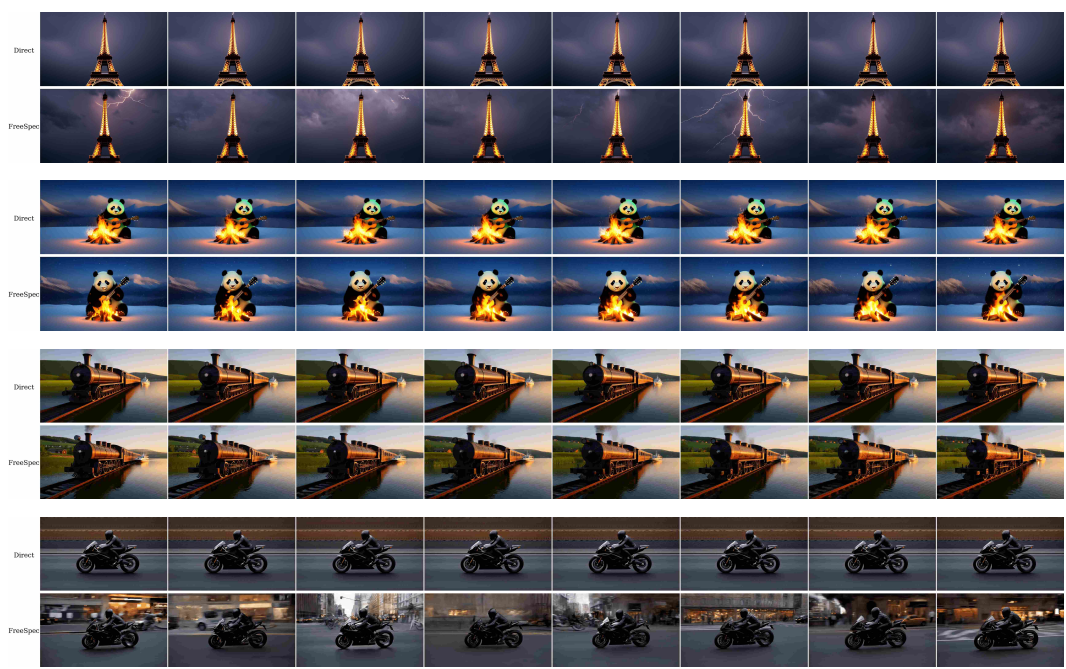}\\[0.6em]
    }
    \caption{
    Additional VBench-Long comparisons between Direct and FreeSpec.
    }
    \label{fig:more_results}
\end{figure*}

\subsection{Additional Results on VBench-Long}

This section reports additional VBench-Long qualitative cases. We compare Direct and FreeSpec to show how singular-spectrum reconstruction affects long-range generation across different prompts.


\end{document}